\documentclass[11pt]{eptcs}
\pdfoutput=1

\usepackage{amsmath} 
\usepackage{amssymb}  
\usepackage{siunitx}
\usepackage{tikz}
\usepackage{wrapfig}
\usepackage{algpseudocode,algorithm,algorithmicx}
\usepackage{tabularx}
\usepackage{dblfloatfix}

\usepackage{outlines}

\usepackage{float}
\usepackage{color}  
\usepackage{xspace}
\usepackage{tabularx}
\usepackage{multirow}
\usepackage{listings}
\usepackage{amsfonts,amsmath}
\usepackage{tabulary}
\newcolumntype{K}[1]{>{\centering\arraybackslash}p{#1}}
\usepackage{enumitem}
\setenumerate[1]{label=\Roman*.}
\setenumerate[2]{label=\Alph*.}
\setenumerate[3]{label=\roman*.}
\setenumerate[4]{label=\alph*.}
\usepackage{xspace}
\usepackage{siunitx}
\usepackage{tikz}
\usetikzlibrary{arrows,calc,decorations.pathmorphing,decorations.markings,backgrounds,decorations,decorations.pathmorphing,positioning,fit,automata,shapes,patterns,decorations.text}
\usepackage{tabularx}
\usepackage{rotating}
\usepackage{outlines}
\usepackage{enumitem}
\usepackage{graphicx} 
\usepackage{color}  
\usepackage{adjustbox}
\usepackage{array}
\usepackage{algpseudocode,algorithm,algorithmicx}
\usepackage{tuncali-dissertation-macros}

\definecolor{codegreen}{rgb}{0,0.6,0}
\definecolor{codegray}{rgb}{0.5,0.5,0.5}
\definecolor{codepurple}{rgb}{0.58,0,0.82}
\definecolor{backcolour}{rgb}{0.95,0.95,0.92}
\lstdefinestyle{pythoncodestyle}{
	backgroundcolor=\color{backcolour},   
	commentstyle=\color{codegreen},
	keywordstyle=\color{magenta},
	numberstyle=\tiny\color{codegray},
	stringstyle=\color{codepurple},
	basicstyle=\footnotesize,
	breakatwhitespace=false,         
	breaklines=true,                 
	captionpos=b,                    
	keepspaces=true,                 
	numbers=left,                    
	numbersep=5pt,                  
	showspaces=false,                
	showstringspaces=false,
	showtabs=false,                  
	tabsize=2
}
\definecolor{commandpromptbackcolour}{rgb}{0.85,0.85,0.85}
\lstdefinestyle{commandpromptstyle}{
	backgroundcolor=\color{commandpromptbackcolour},   
	commentstyle=\color{codegreen},
	keywordstyle=\color{magenta},
	numberstyle=\tiny\color{codegray},
	stringstyle=\color{codepurple},
	basicstyle=\footnotesize,
	breakatwhitespace=false,         
	breaklines=true,                 
	captionpos=b,                    
	keepspaces=true,                 
	numbers=none,                    
	numbersep=5pt,                  
	showspaces=false,                
	showstringspaces=false,
	showtabs=false,                  
	tabsize=2,
	language=command.com
}

\newcommand{\python}{Python\xspace}
\newcommand{\pythonversion}{Python 3.7\xspace}

\providecommand{\publicationstatus}{This article is a modified version of a chapter from Cumhur Erkan Tuncali's Ph.D. Dissertation \cite{tuncali2019dissertation}.}

\title{\LARGE \bf A Tutorial on Sim-ATAV: Simulation-based Adversarial Testing Framework for Autonomous Vehicles}

\author{Cumhur Erkan Tuncali
		\institute{Arizona State University, Tempe, AZ, USA}
	    \email{etuncali@asu.edu}}
	
\def\titlerunning{A Tutorial on Sim-ATAV}
\def\authorrunning{C.E.~Tuncali}

\begin{document}

\maketitle

\begin{abstract}
Testing autonomous vehicles in simulation environments is crucial.
\toolname is an open-source framework developed for experimenting with different test generation techniques in simulation environments for research purposes.
This document provides a tutorial on \toolname with a running example.
\end{abstract}

\section{An Overview of \toolname} 
\toolname \cite{simatav_web,tuncali2018poster,tuncali2018iv} is a tool developed for experimenting with different test generation techniques in simulation environments for research purposes as described in \cite{tuncali2018iv}.
It is mainly developed in \python and it uses the open-source robotics toolbox Webots \cite{webotsweb} for 3D scene generation, vehicle and sensor modeling and simulation.
\toolname can be interfaced with \textit{covering array} generation tools like ACTS \cite{kuhn2013introduction} and with \textit{falsification} tools like \staliro\cite{FainekosSUY12acc}, which is a \matlab toolbox.
\toolname is publicly available as an open-source project \cite{simatav_web}.

\begin{figure}[htp]
	\begin{centering}
		\includegraphics[width=\linewidth,trim={0 11cm 3.5cm 0},clip]{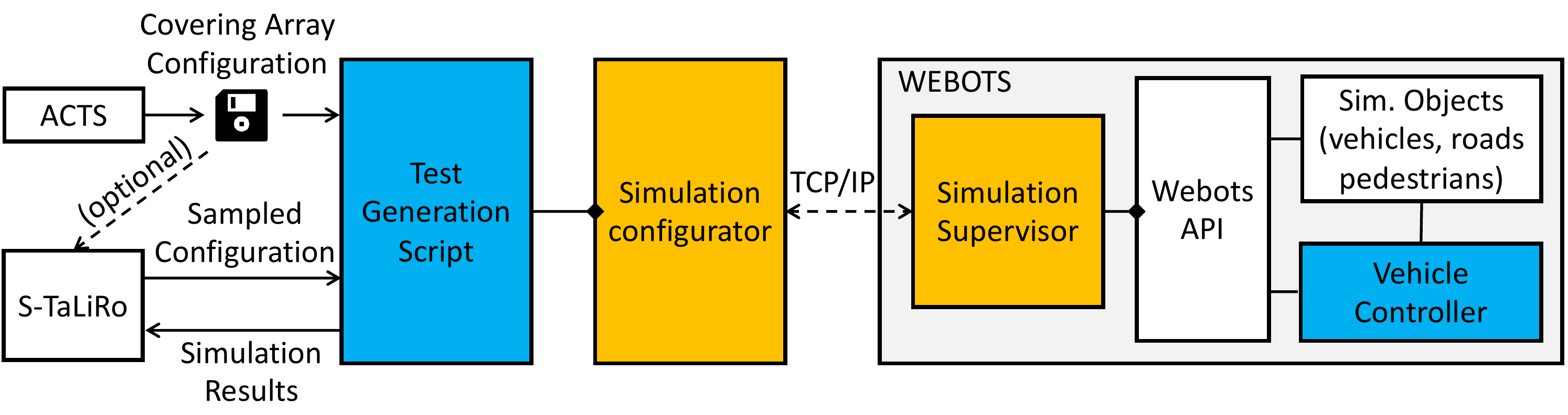}
		\caption{An Overview of Sim-ATAV.}
		\label{fig:simatav_overview}
	\end{centering}
\end{figure}

\figg{\ref{fig:simatav_overview}} provides a high-level overview of the framework.
The main functionality of \toolname is provided by \toolblock{Simulation configurator} and \toolblock{Simulation Supervisor} blocks.
\toolblock{Simulation configurator} block represents the \toolname API for the user script to create a simulation and to receive the results.
\toolblock{Simulation Supervisor} block represents the part of the framework that executes inside the robotics simulation toolbox Webots.
It uses the Webots API to (1) modify the simulation environment, e.g., add/configure vehicles, roads, pedestrians, provide vehicle controllers, (2) execute the simulation, and (3) collect information, e.g., the evolution of vehicle states over the simulation time.
\toolblock{Simulation Supervisor} receives the requested simulation configuration from the \toolblock{Simulation configurator} over socket communication.
When the simulation environment is set up, \toolblock{Simulation Supervisor} requests Webots to start the simulation.
User-provided controllers control the motion of the vehicles and pedestrians.
At the end of the simulation, \toolblock{Simulation Supervisor} sends the collected simulation trace to \toolblock{Simulation configurator}.

External tools like ACTS can be used to generate combinatorial tests with covering arrays.
\toolname provides functions to read covering array test scenarios from \filecall{csv} files.
For falsification, \staliro is used to iteratively sample new configurations until a failure case is detected.
Optionally, initial samples for the configurations can be read by \staliro from the covering array \filecall{csv} files.
The sampled configurations in \staliro are passed as parameters to test generation functions using the \python interface provided in \matlab, \ie by calling \python functions directly inside \matlab.
The test generation functions return the simulation trace back to \matlab (and to \staliro) as return values.
More detail on \textit{falsification} methods is available in \cite{tuncali2018iv,FainekosSUY12acc}.
A running example of test generation is provided in the upcoming sections.

Although they are not essential for the test generation purposes, \toolname also comes with some vehicle controller implementations as well as some basic perception system, sensor fusion, path planning and control algorithms that can be utilized by the user for Ego or Agent vehicle control.

\section{Installation Instructions}
\toolname requires \pythonversion and Webots for basic functionality.  
For some controllers and test generation approaches, there are other requirements like \matlabCopyrighted, \staliro, TensorFlow, \squeezedet etc.  
The framework has been tested in Windows\textsuperscript{\textregistered} 10 with specific versions of the required packages.

Firstly, \toolname should be either downloaded or cloned using a \textit{git} client \footnote{\toolname: https://cpslab.assembla.com/spaces/sim-atav}.
For Windows\textsuperscript{\textregistered}, the preferred installation approach is to use \filecall{setup\_for\_windows.bat}.
Once executed, it will guide the user through the installation process and automate the process as much as possible. 
This script is not advanced and may fail for some systems. 
If the script fails, please try the steps below for a manual installation.

Below are the steps for the manual installation.
All the paths are given relative to the root folder for the \toolname distribution.
In case any problems are experienced during the installation of the packages, most of the packages can also be found in Christoph Gohlke's website\footnote{Christoph Gohlke's website: https://www.lfd.uci.edu/~gohlke/pythonlibs/}:  
\begin{itemize}

\item[1.] Install \pythonversion-64 Bit  
\item[2.] Install Webots r2019a. 
\item[3.] (optional) If the system has CUDA-enabled GPU and it will be utilized for an increased performance:  
\begin{itemize}
\item[a.] Install CUDA Toolkit 10.0  
\item[b.] Install CUDNN 7.3.1  
\end{itemize}
\item[4.] Download \filecall{Python\_Dependencies} from http://www.public.asu.edu/~etuncali/downloads/ and unzip it next to this installation script. 
The \python wheel (\filecall{.whl}) files should be directly under \filecall{./Python\_Dependencies/} 
\item[5.] Install commonly used \python packages:  
\begin{itemize}
\item[a.] Install Numpy+MKL 1.14.6 either from \filecall{Python\_Dependencies} folder or from Christoph Gohlke's website:

\begin{lstlisting}[style=commandpromptstyle]
pip3 install --upgrade 	Python_Dependencies/numpy-1.14.6+mkl-cp37-cp37m-win_amd64.whl
\end{lstlisting}

\item[b.] Install scipy 1.2.0:
\begin{lstlisting}[style=commandpromptstyle]
pip3 install scipy==1.2.0
\end{lstlisting}

\item[c.] Install scikit-learn:
\begin{lstlisting}[style=commandpromptstyle]
pip3 install scikit-learn
\end{lstlisting}

\item[d.] Install pandas:
\begin{lstlisting}[style=commandpromptstyle]
pip3 install pandas
\end{lstlisting}

\item[e.] Install Absl Py:
\begin{lstlisting}[style=commandpromptstyle]
pip3 install absl-py
\end{lstlisting}

\item[f.] Install matplotlib:
\begin{lstlisting}[style=commandpromptstyle]
pip3 install matplotlib
\end{lstlisting}

\item[g.] Install pykalman:
\begin{lstlisting}[style=commandpromptstyle]
pip3 install pykalman
\end{lstlisting}

\item[h.] Install Shapely:
\begin{lstlisting}[style=commandpromptstyle]
pip3 install Shapely
\end{lstlisting}
\begin{itemize}
\item[!] If any problems are experienced during the installation of Shapely, it can be installed from \filecall{Python\_Dependencies} folder:  
\begin{lstlisting}[style=commandpromptstyle]
pip3 install Python_Dependencies/Shapely-1.6.4.post1-cp37-cp37m-win_amd64.whl
\end{lstlisting}
\end{itemize}

\item[i.] Install dubins:
\begin{lstlisting}[style=commandpromptstyle]
pip3 install dubins
\end{lstlisting}

\begin{itemize}
\item[!] If any problems are experienced when installing pydubins:  

One option is to go into the folder \filecall{Python\_Dependencies/pydubins} and execute \filecall{python\ setup.py install}.
Another option is the following:  
\begin{itemize}
\item[(i)] Download pydubins from github.com/AndrewWalker/pydubins.  
\item[(ii)] Do the following changes in \filecall{dubins/src/dubins.c}:  
\begin{lstlisting}[language=c,numbers=none,backgroundcolor=\color{gray!10}]
#ifndef M_PI
#define M_PI 3.14159265358979323846  
#endif  
\end{lstlisting}

\item[(iii)] Call \filecall{python\ setup.py install} inside \filecall{pydubins} folder.  
\end{itemize}
\end{itemize}
\end{itemize}

\item[6.] For controllers with DNN (Deep Neural Network) object detection:  

\textit{! Currently, \pythonversion support for Tensorflow is provided by a 3rd party (only for Windows). 
Installation wheels are provided under Python\_Dependencies folder. } 

Check if the system CPU supports AVX2 (for increased performance) 
\footnote{A list of CPUs with AVX2 is available at:\\ https://en.wikipedia.org/wiki/Advanced\_Vector\_Extensions\#CPUs\_with\_AVX2}.
\begin{itemize}

\item[a.] If the system GPU has CUDA cores, CUDNN is installed and the system CPU supports AVX 2: Install Tensorflow-gpu with AVX2 support.  
\begin{lstlisting}[style=commandpromptstyle]
pip3 install --upgrade Python_Dependencies/tensorflow_gpu-1.12.0-cp37-cp37m-win_amd64.whl
\end{lstlisting}

\item[b.] If the system GPU has CUDA cores, CUDNN is installed and the system CPU does NOT support AVX 2: Install Tensorflow-gpu without AVX2 support.  
\begin{lstlisting}[style=commandpromptstyle]
pip3 install --upgrade Python_Dependencies/sse2/tensorflow_gpu-1.12.0-cp37-cp37m-win_amd64.whl
\end{lstlisting}

\item[c.] If the system GPU does not have CUDA cores or CUDNN is not installed, and the system CPU supports AVX 2: Install Tensorflow with AVX2 support.  
\begin{lstlisting}[style=commandpromptstyle]
pip3 install --user --upgrade Python_Dependencies/tensorflow-1.12.0-cp37-cp37m-win_amd64.whl
\end{lstlisting}

\item[d.] If the system GPU does not have CUDA cores or CUDNN is not installed, and the system CPU does NOT support AVX 2: Install Tensorflow without AVX2 support.  
\begin{lstlisting}[style=commandpromptstyle]
pip3 install --user --upgrade Python_Dependencies/sse2/tensorflow-1.12.0-cp37-cp37m-win_amd64.whl
\end{lstlisting}

\end{itemize}
\item[7.] Install Python Dependencies of SqueezeDet (if the existing controllers that use \squeezedet will be used). 
There is no need to install \squeezedet separately, as it is provided in the framework.  
\begin{itemize}
\item[a.] Install joblib:
\begin{lstlisting}[style=commandpromptstyle]
pip3 install --upgrade joblib
\end{lstlisting}

\item[b.] Install opencv:
\begin{lstlisting}[style=commandpromptstyle]
pip3 install --upgrade opencv-contrib-python
\end{lstlisting}

\item[c.] Install pillow:
\begin{lstlisting}[style=commandpromptstyle]
pip3 install --upgrade Pillow
\end{lstlisting}

\item[d.] Install easydict:
\begin{lstlisting}[style=commandpromptstyle]
pip3 install --upgrade easydict==1.7
\end{lstlisting}

! If any problems are experienced while installing easydict, try the following:  
\begin{lstlisting}[style=commandpromptstyle]
cd Python_Dependencies/easydict-1.7\
python setup.py install
cd ../..
\end{lstlisting}

\end{itemize}

\item[8.] To design Covering Array Tests:
Please request a copy and install ACTS from NIST\footnote{ACTS tool can be requested from\\
	 https://csrc.nist.gov/projects/automated-combinatorial-testing-for-software/downloadable-tools}.
\item[9.] To do robustness-guided falsification, \matlabCopyrighted and \staliro are needed:  
\begin{itemize}
\item[a.] Install \matlab from Mathworks(tested with r2017b).  
\item[b.] Install \staliro \footnote{\staliro is available at https://sites.google.com/a/asu.edu/s-taliro/s-taliro}.  
\end{itemize}

\item[10.] After the installation is finished, the \python package wheel files that are under the folder named \filecall{Python\_Dependencies} can be deleted to save some disk space.  
\end{itemize}

\paragraph{Setting to Utilize GPU:}
If the system has a CUDA-enabled GPU, and CUDA Toolkit, CUDNN are installed, the variable \filecall{has\_gpu} should be set to \filecall{True} in the following file to make the experiments use the system GPU for \squeezedet: 

\filecall{Sim\_ATAV/classifier/classifier\_interface/gpu\_check.py}.

\section{Reference Manual}
\subsection{Simulation Entities}
\toolname starts building a testing scenario from an existing Webots world file provided by the user, which we will call as \textit{base world file}.
The user has the option to have all or some of the simulation entities, \ie roads, vehicles, etc., saved in the base world file before the test generation time.
The user can use the functionality provided by \toolname to add more simulation entities to the world at the time of test generation.
This functionality is especially useful when the search space for the tests contain some parameters of the simulation entities such as road width, number of lanes, positions of the vehicles.

This section describes the most commonly used simulation entities that can be programmatically added into the simulation world at the time of test generation.
Note that \toolname may not provide the functionality to add all possible types of simulation objects that are supported by Webots.
In this section, the class names and properties with their default values for the simulation entities supported by \toolname are provided.
\toolblock{Test Generation Script} which is developed by the user typically creates instances of required simulation entities and uses the provided functions to add those entities into the test scenario.
For modifying the simulation environment beyond the capabilities of \toolname, the user can do the changes manually and save in the base world file or modify the source code of \toolname to add or change the capabilities as needed.
For a deeper understanding of the possibilities, the reader is advised to get familiar with the Webots simulation environment and details of available simulation entities \cite{webotsweb}.

The class properties for the simulation entities are provided in the tables below.
The first row of each table gives the class name.
Other rows start with the property name, gives the default value of the property (the value used if not explicitly changed by the user), and a brief description of the property.
For each class, a related source code snippet from a running example is provided.
The original source code for the running example can be found as \filecall{tests/tutorial\_example\_1.py} in the \toolname distribution.

\begin{figure*}[htp]
	\begin{centering}
		\includegraphics[width=0.95\linewidth]{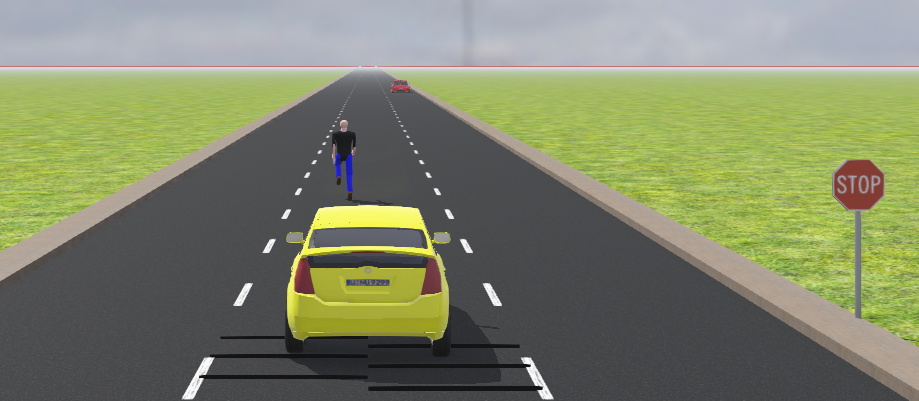}
		\caption{A view from the generated scenario for the running example.}
		\label{fig:scenario_view}
	\end{centering}
\end{figure*}

\figg{\ref{fig:scenario_view}} is an image from the scenario described in the running example.
A yellow Ego vehicle is behind a pedestrian walking in the middle of a 3-lane road, and an agent vehicle is approaching from the opposite direction in the next lane.
There is a bumpy road surface for a short distance, and a stop sign placed on the right side of the road.
This is only a simple example to illustrate how to use \toolname.

\paragraph{General information:}
In Webots, all simulation entities are kept in a simulation tree and they can be easily accessed using their \textbf{DEF\_NAME} property.
\textbf{Position} fields are 3D arrays as $[x, y, z]$, keeping the value of the position at each axis in Webots coordinate system, where $y$ is typically the vertical axis to the ground (in practice, this depends on the Webots world file provided by the user).
\textbf{Rotation} fields are 4D arrays as $[x, y, z, \theta]$, where $x,y,z$ represents a rotation vector and $\theta$ represents the rotation around this vector in clockwise direction. 

\subsubsection{Simulation Environment}
Once an object is created for a simulation entity, the user can utilize the corresponding function provided by \toolname to add the entity to the scenario.
An easier alternative is to utilize the \textbf{SimEnvironment} class provided by \toolname.
An object of this class can be populated with the necessary simulation entities and passed to the simulation environment with a single function call.
\tbl{\ref{tab:simenvironmentclass}} summarizes the \textbf{SimEnvironment} class.

\begin{table}[H]
	\begin{center}
		\caption{Simulation Environment Class.}
		\label{tab:simenvironmentclass}
		\begin{tabularx}{\textwidth}{llX}
			\hline
			\multicolumn{3}{l}{Class Name: \textbf{SimEnvironment}}\\\hline
			\textbf{Property} & \textbf{Default} & \textbf{Description}\\\hline
			fog & None & Keeps fog object.\\
			heart\_beat\_config & None & The heartbeat configuration.\\
			view\_follow\_config & None & The viewpoint configuration.\\
			ego\_vehicles\_list & [] & The list of Ego vehicles.\\
			agent\_vehicles\_list & [] & The list of agent vehicles.\\
		    pedestrians\_list & [] & The list of pedestrians.\\
			road\_list & [] & The list of roads.\\
			road\_disturbances\_list & [] & The list of road disturbances.\\
			generic\_sim\_objects\_list & [] & The list of generic simulation objects.\\
			control\_params\_list & [] & The list of controller parameters that will be set in the run time.\\
			initial\_state\_config\_list & [] & The list of initial state configurations.\\
			data\_log\_description\_list & [] & The list of data log descriptions.\\
			data\_log\_period\_ms & None & Data log period (ms).\\\hline
		\end{tabularx}
	\end{center}
\end{table}

\begin{exmp}
For the running example, we start with an empty simulation environment.
Listing \ref{lst:simenvironmentclass} creates an empty \textbf{SimEnvironment} object that will later keep the required simulation entities.

\begin{lstlisting}[language=Python, caption=Source code for creating a simulation environment., label=lst:simenvironmentclass]
from Sim_ATAV.simulation_configurator.sim_environment \
    import SimEnvironment
    
sim_environment = SimEnvironment()
\end{lstlisting}
\end{exmp}

\newpage
\subsubsection{Road}
A user-configurable road structure to use in the simulation environment is given in \tbl{\ref{tab:roadclass}}.

\begin{table}[H]
	\begin{center}
		\caption{Simulation Entity: Road.}
		\label{tab:roadclass}
		\begin{tabularx}{\textwidth}{llX}
			\hline
			\multicolumn{3}{l}{Class Name: \textbf{WebotsRoad}}\\\hline
			\textbf{Property} & \textbf{Default} & \textbf{Description}\\\hline
			def\_name & ``STRROAD'' & Name as it appears in the simulation tree. \\
			road\_type & ``StraightRoadSegment'' & Road type name as used by Webots. \\
			rotation & [0, 1, 0, math.pi/2] & Rotation of the road \\
			position & [0, 0.02, 0] & Starting position. \\
			number\_of\_lanes & 2 & Number of lanes.\\
			width & number\_of\_lanes * 3.5 & Road width (m). \\
			length & 1000 & Road length (m).\\
			\hline
		\end{tabularx}
	\end{center}
\end{table}

\begin{exmp}
Listing \ref{lst:roadclass} provides a source code snippet that creates a 3-lane straight road segment lying between 1000m and -1000m along the x-axis.

\begin{lstlisting}[language=Python, caption=Source code for creating a road., label=lst:roadclass]
from Sim_ATAV.simulation_control.webots_road import WebotsRoad

road = WebotsRoad(number_of_lanes=3)
road.rotation = [0, 1, 0, -math.pi / 2]
road.position = [1000, 0.02, 0]
road.length = 2000.0

# Add the road into simulation environment object:
sim_environment.road_list.append(road)
\end{lstlisting}
\end{exmp}

\pagebreak
\subsubsection{Vehicle}
A user-configurable vehicle class to use in the simulation environment in \tbl{\ref{tab:vehicleclass}}.

\begin{table}[H]
	\begin{center}
		\caption{Simulation Entity: Vehicle.}
		\label{tab:vehicleclass}
		\begin{tabularx}{\textwidth}{llX}
			\hline
			\multicolumn{3}{l}{Class Name: \textbf{WebotsVehicle}}\\\hline
			\textbf{Property} & \textbf{Default} & \textbf{Description}\\\hline
			def\_name & ``'' & Name as it appears in Webots simulation tree. \\
			vhc\_id & 0 & Integer ID for referencing to the vehicle. \\
			vehicle\_model & ``AckermannVehicle'' & Vehicle model can be any model name available in Webots. \\
			rotation & [0, 1, 0, 0] & Rotation of the object. \\
			current\_position & [0, 0.3, 0] & x,y,z values of the position. \\
			color & [1, 1, 1] & R,G,B values of the color in the range [0,1]. \\
			controller & ``void'' & Name of the vehicle controller.\\
			is\_controller\_name\_absolute & False & Indicates where to find the vehicle controller. Find the details below. \\
			vehicle\_parameters & [] & Additional parameters for the vehicle object.\\
			controller\_parameters & [] & Parameters that will be sent to the vehicle controller.\\
			controller\_arguments & True & Arguments passed to the vehicle controller executable.\\
			sensor\_array & [] & List of sensors on the vehicle (\textbf{WebotsSensor} objects).\\\hline
		\end{tabularx}
	\end{center}
\end{table}

The \filecall{vehicle\_model} field of a vehicle object should match with the models available in Webots (or any custom models added to Webots by the user). 
Webots r2019a version provides vehicle model options \textit{ToyotaPrius, CitroenCZero, BmwX5, RangeRoverSportSVR, LincolnMKZ, TeslaModel3} as well as truck, motorcycle and tractor models.
When \filecall{is\_controller\_name\_absolute} is set to true, Webots will load the given controller from \filecall{Webots\_Projects/controllers} folder, otherwise, it will load the controller named \filecall{vehicle\_controller} which is located under the same folder but will take the controller name as an argument.

\begin{exmp}
We can now create vehicles and place them on the road that was created above.
Listing \ref{lst:vehicleclass} provides an example source code snippet that creates an Ego vehicle at the position $x=20m$, $y=0$, and an agent vehicle at the position $x=300m$, $y=3.5m$, vehicles facing toward each other.
The controllers for the vehicles are set and the controller arguments are provided.
The arguments accepted are controller-specific.
The vehicle controllers used in this example can be found under the folder \filecall{Webots\_Projects/controllers}.

\begin{lstlisting}[language=Python, caption=Source code for creating a vehicle., label=lst:vehicleclass]
from Sim_ATAV.simulation_control.webots_vehicle import WebotsVehicle

# Ego vehicle
ego_x_pos = 20.0 # Setting the x position of the Ego vehicle in a variable.

vhc_obj = WebotsVehicle()
vhc_obj.current_position = [ego_x_pos, 0.35, 0.0]
vhc_obj.current_orientation = math.pi/2
vhc_obj.rotation = [0.0, 1.0, 0.0, vhc_obj.current_orientation]
vhc_obj.vhc_id = 1
vhc_obj.color = [1.0, 1.0, 0.0]
vhc_obj.set_vehicle_model('ToyotaPrius')
# Name of our controller python file is 'automated_driving_with_fusion2':
vhc_obj.controller = 'automated_driving_with_fusion2'
# Controller will be found directly under controllers folder:
vhc_obj.is_controller_name_absolute = True
# Below is a list of arguments specific to this controller.
# For reference, the arguments are: car_model, target_speed_kmh, target_lat_pos, 
# self_vhc_id, slow_at_intersection, has_gpu, processor_id
vhc_obj.controller_arguments.append('Toyota')
vhc_obj.controller_arguments.append('70.0')
vhc_obj.controller_arguments.append('0.0')
vhc_obj.controller_arguments.append('1')
vhc_obj.controller_arguments.append('True')
vhc_obj.controller_arguments.append('False')
vhc_obj.controller_arguments.append('0')

# Agent vehicle:
vhc_obj2 = WebotsVehicle()
vhc_obj2.current_position = [300.0, 0.35, 3.5]
vhc_obj2.current_orientation = 0.0
vhc_obj2.rotation = [0.0, 1.0, 0.0, -math.pi/2]
vhc_obj2.vhc_id = 2
vhc_obj2.set_vehicle_model('TeslaModel3')
vhc_obj2.color = [1.0, 0.0,  0.0]
vhc_obj2.controller = 'path_and_speed_follower'
vhc_obj2.controller_arguments.append('20.0')
vhc_obj2.controller_arguments.append('True')
vhc_obj2.controller_arguments.append('3.5')
vhc_obj2.controller_arguments.append('2')
vhc_obj2.controller_arguments.append('False')
vhc_obj2.controller_arguments.append('False')

# Here, we don't save the vehicles into simulation environment yet
# because we will add some sensors on the vehicles below.
\end{lstlisting}
\end{exmp}

\subsubsection{Sensor}
\tbl{\ref{tab:sensorclass}} provides an overview of the \textbf{WebotsSensor} class that can be used to describe a sensor.
Webots vehicle models have specific sensor slots on the vehicles.
These are typically \filecall{TOP, CENTER, FRONT, RIGHT, LEFT}.
The property \filecall{sensor\_type} accepts the type of the sensor which should match the type used by Webots.
The \textit{translation} field of the sensor can be used to place the sensor to a different position relative to its corresponding sensor slot.
As sensors can vary a lot in terms or parameters, \filecall{sensor\_fields} property is provided to accept names and values of the desired parameters as a list of \textbf{WebotsSensorField} objects for flexible configuration of sensors.

\begin{table}[H]
	\begin{center}
		\caption{Simulation Entity: Sensor.}
		\label{tab:sensorclass}
		\begin{tabularx}{\textwidth}{llX}
			\hline
			\multicolumn{3}{l}{Class Name: \textbf{WebotsSensor}}\\\hline
			\textbf{Property} & \textbf{Default} & \textbf{Description}\\\hline
			sensor\_type & ``'' & Type of the sensor as defined in Webots. \\
			sensor\_location & FRONT & Sensor slot enumeration.\\
			& & $<$FRONT, CENTER, LEFT, RIGHT, TOP$>$\\
			sensor\_fields & [] & List of \textbf{WebotsSensorField} objects to customize the sensor. \\\hline
			\multicolumn{3}{l}{Class Name: \textbf{WebotsSensorField}}\\\hline
			field\_name & ``'' & Name of the field that will be set. \\
			field\_val & ``'' & Value of the field.\\\hline
		\end{tabularx}
	\end{center}
\end{table}

\begin{exmp}
An example source code of adding sensors to vehicles is provided in Listing \ref{lst:sensorclass}.
In this example, we add a \textit{compass} and a \textit{GPS} device that are used by the controllers for path following.
The \textit{receiver} device added to the vehicles is used by the controllers to receive new commands from \toolblock{Simulation Supervisor}.
We add the receivers because we will later need them to update target paths of the vehicles.
Note that although the necessary infrastructure for this approach is provided by \toolname, it is implementation specific and not mandated.
In this example, we also add a \textit{radar} device to Ego vehicle for collision avoidance.

\begin{lstlisting}[language=Python, caption=Source code for adding sensor to a vehicle., label=lst:sensorclass]
from Sim_ATAV.simulation_control.webots_sensor import WebotsSensor

# Add a radio receiver to the center sensor slot
# with the name field set to 'receiver'. 
# This is optional and will be used to communicate with the controller at the run-time.
vhc_obj.sensor_array.append(WebotsSensor())
vhc_obj.sensor_array[-1].sensor_location = WebotsSensor.CENTER
vhc_obj.sensor_array[-1].sensor_type = 'Receiver'
vhc_obj.sensor_array[-1].add_sensor_field('name', '"receiver"')

# Add a compass to the center slot with the name field set to 'compass'.
vhc_obj.sensor_array.append(WebotsSensor())
vhc_obj.sensor_array[-1].sensor_location = WebotsSensor.CENTER
vhc_obj.sensor_array[-1].sensor_type = 'Compass'
vhc_obj.sensor_array[-1].add_sensor_field('name', '"compass"')

# Add a GPS receiver to the center sensor slot.
vhc_obj.sensor_array.append(WebotsSensor())
vhc_obj.sensor_array[-1].sensor_location = WebotsSensor.CENTER
vhc_obj.sensor_array[-1].sensor_type = 'GPS'

# Add a Radar to the front sensor slot with the name field set to 'radar'.
vhc_obj.sensor_array.append(WebotsSensor())
vhc_obj.sensor_array[-1].sensor_type = 'Radar'
vhc_obj.sensor_array[-1].sensor_location = WebotsSensor.FRONT
vhc_obj.sensor_array[-1].add_sensor_field('name', '"radar"')

# Finally, add the vehicle to the simulation environment as an Ego vehicle.
sim_environment.ego_vehicles_list.append(vhc_obj)

# Similar for the agent vehicle:
vhc_obj2.sensor_array.append(WebotsSensor())
vhc_obj2.sensor_array[-1].sensor_location = WebotsSensor.CENTER
vhc_obj2.sensor_array[-1].sensor_type = 'Receiver'
vhc_obj2.sensor_array[-1].add_sensor_field('name', '"receiver"')

vhc_obj2.sensor_array.append(WebotsSensor())
vhc_obj2.sensor_array[-1].sensor_location = WebotsSensor.CENTER
vhc_obj2.sensor_array[-1].sensor_type = 'Compass'
vhc_obj2.sensor_array[-1].add_sensor_field('name', '"compass"')

vhc_obj2.sensor_array.append(WebotsSensor())
vhc_obj2.sensor_array[-1].sensor_location = WebotsSensor.CENTER
vhc_obj2.sensor_array[-1].sensor_type = 'GPS'

# Add the agent vehicle to the simulation environment
sim_environment.agent_vehicles_list.append(vhc_obj2)
\end{lstlisting}
\end{exmp}

\subsubsection{Pedestrian}
The user-configurable pedestrian class, \textbf{WebotsPedestrian}, to use in the simulation environment is described in \tbl{\ref{tab:pedestrianclass}}.
Target speed and target path (trajectory) of the pedestrian are automatically passed as arguments to the given controller.

\begin{table}[H]
	\begin{center}
		\caption{Simulation Entity: Pedestrian.}
		\label{tab:pedestrianclass}
		\begin{tabularx}{\textwidth}{llX}
			\hline
			\multicolumn{3}{l}{Class Name: \textbf{WebotsPedestrian}}\\\hline
			\textbf{Property} & \textbf{Default} & \textbf{Description}\\\hline
			def\_name & ``PEDESTRIAN'' & Name as it appears in Webots simulation tree. \\
			ped\_id & 0 & Integer ID for referencing to the pedestrian. \\
			rotation & [0, 1, 0, math.pi/2.0] & Rotation of the object. \\
			current\_position & [0, 0, 0] & x,y,z values of the position. \\
			shirt\_color & [0.25, 0.55, 0.2] & R,G,B values of the shirt color in the range [0, 1]. \\
			pants\_color & [0.24, 0.25, 0.5] & R,G,B values of the pants color in the range [0, 1]. \\
			shoes\_color & [0.28, 0.15, 0.06] & R,G,B values of the shoes color in the range [0, 1]. \\
			controller & ``void'' & Name of the pedestrian controller.\\
			target\_speed & 0.0 & Walking speed of the pedestrian. \\
			trajectory & [] & Walking path of the pedestrian.\\\hline
		\end{tabularx}
	\end{center}
\end{table}

\newpage

\begin{exmp}
	
Listing \ref{lst:pedestrianclass} provides an example code snippet to create a pedestrian object, and provide a target speed and a path to define the motion of the pedestrian.

\begin{lstlisting}[language=Python, caption=Source code for creating a pedestrian., label=lst:pedestrianclass]
from Sim_ATAV.simulation_control.webots_pedestrian import WebotsPedestrian

pedestrian_speed = 3.0 # Setting the pedestrian walking speed in a variable.

pedestrian = WebotsPedestrian()
pedestrian.ped_id = 1
pedestrian.current_position = [50.0, 1.3, 0.0]
pedestrian.shirt_color = [0.0, 0.0, 0.0]
pedestrian.pants_color = [0.0, 0.0, 1.0]
pedestrian.target_speed = pedestrian_speed
# Pedestrian trajectory as a list of x1, y1, x2, y2, ...
pedestrian.trajectory = [50.0, 0.0, 80.0, -3.0, 200.0, 0.0]
pedestrian.controller = 'pedestrian_control'

# Add the pedestrian into the simulation environment.
sim_environment.pedestrians_list.append(pedestrian)
\end{lstlisting}
\end{exmp}

\newpage
\subsubsection{Fog}
User-configurable fog class to use in the simulation environment is summarized in \tbl{\ref{tab:fogclass}}.

\begin{table}[H]
	\begin{center}
		\caption{Simulation Entity: Fog.}
		\label{tab:fogclass}
		\begin{tabularx}{\textwidth}{llX}
			\hline
			\multicolumn{3}{l}{Class Name: \textbf{WebotsFog}}\\\hline
			\textbf{Property} & \textbf{Default} & \textbf{Description}\\\hline
			def\_name & ``FOG'' & Name as it appears in Webots simulation tree. \\
			fog\_type & ``LINEAR'' & Defines the type of the fog gradient. \\
			color & [0.93, 0.96, 1.0] & R,G,B values of the fog color in the range [0, 1]. \\
			visibility\_range & 1000 & Visibility range of the fog (m).\\\hline
		\end{tabularx}
	\end{center}
\end{table}

\begin{exmp}
	
No camera is involved in this scenario, hence fog will not impact the performance of the controller.
However, an example source code snippet is provided in Listing \ref{lst:fogclass} for reference.

\begin{lstlisting}[language=Python, caption=Source code for creating fog., label=lst:fogclass]
from Sim_ATAV.simulation_control.webots_fog import WebotsFog

# Creating fog with 700m visibility and adding it to the simulation environment.
sim_environment.fog = WebotsFog()
sim_environment.fog.visibility_range = 700.0
\end{lstlisting}
\end{exmp}

\subsubsection{Road Disturbance}
Road disturbance objects are solid triangular objects placed on the road surface to emulate a bumpy road surface.
\textbf{WebotsRoadDisturbance}, which is described in \tbl{\ref{tab:roaddisturbanceclass}}, contains the properties to describe how the solid objects will be placed to create a bumpy road surface.

\begin{table}[H]
	\begin{center}
		\caption{Simulation Entity: Road Disturbance Object.}
		\label{tab:roaddisturbanceclass}
		\begin{tabularx}{\textwidth}{llX}
			\hline
			\multicolumn{3}{l}{Class Name: \textbf{WebotsRoadDisturbance}}\\\hline
			\textbf{Property} & \textbf{Default} & \textbf{Description}\\\hline
			disturbance\_id & 1 & Object ID for later reference. \\
			disturbance\_type & INTERLEAVED & Enumerated type of the disturbance.\\
			& & $<$INTERLEAVED, \\
			& & FULL\_LANE\_LENGTH, \\
			& & ONLY\_LEFT, ONLY\_RIGHT$>$\\
			rotation & [0, 1, 0, 0] & Rotation of the object. \\
			position & [0, 0, 0] & x,y,z values of the position. \\
			length & 100 & Length of the bumpy road surface (m).\\
			width & 3.5 & Width of the corresponding lane.\\
			height & 0.06 & Height of the disturbance (m).\\
			surface\_height & 0.02 & Height of the corresponding road surface (m).\\
			inter\_object\_spacing & 1.0 & Distance between repeating solid objects on the road (m).\\
			\hline
		\end{tabularx}
	\end{center}
\end{table}

\begin{exmp}
	
An example road disturbance object creation is given in Listing \ref{lst:roaddisturbanceclass}.

\begin{lstlisting}[language=Python, caption=Source code for creating road disturbance., label=lst:roaddisturbanceclass]
from Sim_ATAV.simulation_control.webots_road_disturbance import WebotsRoadDisturbance

# Create bumpy road for 3m where there are road disturbances on both side of the lane
# of height 4cm, each separated with 0.5m.
road_disturbance = WebotsRoadDisturbance()
road_disturbance.disturbance_type = WebotsRoadDisturbance.TRIANGLE_DOUBLE_SIDED  #i.e., INTERLEAVED
road_disturbance.rotation = [0, 1, 0, -math.pi / 2.0]  # Same as the road
road_disturbance.position = [40, 0, 0]
road_disturbance.width = 3.5
road_disturbance.length = 3
road_disturbance.height = 0.04
road_disturbance.inter_object_spacing = 0.5

# Add road disturbance into the simulation environment object.
sim_environment.road_disturbances_list.append(road_disturbance)
\end{lstlisting}
\end{exmp}

\subsubsection{Generic Simulation Object}
Generic simulation object is for adding any type of object into the simulation which is not covered above.
For these objects, there are no checks performed or there are no limitations on the field values.
The user can manually create any possible Webots object by setting all of its non-default field values.

\begin{table}[H]
	\begin{center}
		\caption{Simulation Entity: Generic Simulation Object.}
		\label{tab:simobjectclass}
		\begin{tabularx}{\textwidth}{llX}
			\hline
			\multicolumn{3}{l}{Class Name: \textbf{WebotsSimObject}}\\\hline
			\textbf{Property} & \textbf{Default} & \textbf{Description}\\\hline
			def\_name & ``'' & Name as it appears in Webots simulation tree. \\
			object\_name & ``Tree'' & Type name of the object. Must be same as the name used by Webots.\\
			object\_parameters & [] & List of (field name, field value) tuples as strings. Names must be same as the field names used by Webots. \\\hline
		\end{tabularx}
	\end{center}
\end{table}

\begin{exmp}
In Listing \ref{lst:simobjectclass}, although it is not expected to have an impact on the controller performance, a Stop Sign object is created as a generic simulation object example for reference.
\begin{lstlisting}[language=Python, caption=Source code for adding a stop sign to the simulation., label=lst:simobjectclass]
from Sim_ATAV.simulation_control.webots_sim_object import WebotsSimObject

sim_obj = WebotsSimObject()
sim_obj.object_name = 'StopSign'  # The name of the object as defined in Webots.
# The field names and format as they are used by Webots.
sim_obj.object_parameters.append(('translation', '40 0 6'))
sim_obj.object_parameters.append(('rotation', '0 1 0 1.5708'))

# Add the stop sign as a generic item into the simulation environment object.
sim_environment.generic_sim_objects_list.append(sim_obj)
\end{lstlisting}
\end{exmp}

\subsection{Configuring the Simulation Execution}

\subsubsection{Additional Controller Parameters}
Depending on the application and implementation details, controller parameters can be directly given to the controllers or they can be sent in the run-time.
To emulate runtime inputs, such as human commands, \toolname can transmit controller commands over virtual radio communication.
The controller should be able to read those commands from a radio receiver and a \textit{receiver} object should be added to one of the sensor slots of the vehicles.
This is an optional approach and the user is free to use other approaches such as reading data from a file, using socket communications etc.

\begin{table}[H]
	\begin{center}
		\caption{Controller parameters.}
		\label{tab:controllerparamsclass}
		\begin{tabularx}{\textwidth}{llX}
			\hline
			\multicolumn{3}{l}{Class Name: \textbf{WebotsControllerParameter}}\\\hline
			\textbf{Property} & \textbf{Default} & \textbf{Description}\\\hline
			vehicle\_id & None & ID of the corresponding vehicle.\\
			parameter\_name & ``'' & Name of the parameter as string.\\
			parameter\_data & [] & Parameter data.\\\hline
		\end{tabularx}
	\end{center}
\end{table}

\begin{exmp}
An example controller parameter creation.
\begin{lstlisting}[language=Python, caption=Source code for creating controller parameters.]
from Sim_ATAV.simulation_control.webots_controller_parameter \
  import WebotsControllerParameter

# ----- Controller Parameters:
# Ego Target Path:
target_pos_list = [[-1000.0, 0.0], [1000.0, 0.0]]

# Add each target position as a controller parameter for Ego vehicle.
for target_pos in target_pos_list:
  sim_environment.controller_params_list.append(
    WebotsControllerParameter(vehicle_id=1,
      parameter_name='target_position',
      parameter_data=target_pos))

# Agent Target Path:
target_pos_list = [[1000.0, 3.5], [145.0, 3.5],  [110.0, -3.5], [-1000.0, -3.5]]

# Add each target position as a controller parameter for agent vehicle.
for target_pos in target_pos_list:
  sim_environment.controller_params_list.append(
    WebotsControllerParameter(vehicle_id=2,
      parameter_name='target_position',
      parameter_data=target_pos))
\end{lstlisting}
\end{exmp}

\subsubsection{Heartbeat Configuration}
\toolblock{Simulation Supervisor} can periodically report the status of the simulation execution to \toolblock{Simulation Configurator} with heartbeats.
\toolblock{Simulation Configurator} can further modify the simulation environment on the run time by responding to the heartbeats.

\begin{table}[H]
	\begin{center}
		\caption{Heartbeat Configuration.}
		\label{tab:heartbeatconfigclass}
		\begin{tabularx}{\textwidth}{llX}
			\hline
			\multicolumn{3}{l}{Class Name: \textbf{WebotsSimObject}}\\\hline
			\textbf{Property} & \textbf{Default} & \textbf{Description}\\\hline
			sync\_type & NO\_HEART\_BEAT & NO\_HEART\_BEAT: Do not report simulation status. \\
			& & WITHOUT\_SYNC: Report the status and continue simulation. \\
			& & WITH\_SYNC: Wait for new commands after each heartbeat. \\
			period\_ms & 10 & Period of the simulation status reporting.\\\hline
		\end{tabularx}
	\end{center}
\end{table}

\begin{exmp}
An example heartbeat configuration.
\begin{lstlisting}[language=Python, caption=Source code for creating a heartbeat configuration.]
from Sim_ATAV.simulation_control.heart_beat import HeartBeatConfig

# Create a heartbeat configuration that will make Simulation Supervisor report 
# simulation status at every 2s and continue execution without waiting for a new 
# command.
sim_environment.heart_beat_config = \
  HeartBeatConfig(sync_type=HeartBeatConfig.WITHOUT_SYNC, period_ms=2000)
\end{lstlisting}
\end{exmp}

\subsubsection{Data Log Item}
Data log items are the states that will be collected into the simulation trace by \toolblock{Simulation Supervisor}.
\begin{table}[H]
	\begin{center}
		\caption{Data Item Description.}
		\label{tab:itemdescriptionclass}
		\begin{tabularx}{\textwidth}{p{4cm}p{3cm}X}
			\hline
			\multicolumn{3}{l}{Class Name: \textbf{ItemDescription}}\\\hline
			\textbf{Property} & \textbf{Default} & \textbf{Description}\\\hline
			item\_type & None & Type of the corresponding simulation entity. \\
			& & $<$TIME, VEHICLE, PEDESTRIAN$>$\\
			item\_index & None & Index of the corresponding simulation entity.\\
			item\_state\_index & None & Index of the state that will be recorded.\\\hline
		\end{tabularx}
	\end{center}
\end{table}

\begin{exmp}
An example list of data log items for simulation trajectory generation.
\begin{lstlisting}[language=Python, caption=Source code for creating data log items.]
 # ----- Data Log Configurations:
 # First entry in the simulation trace will be the simulation time:
sim_environment.data_log_description_list.append(
  ItemDescription(item_type=ItemDescription.ITEM_TYPE_TIME, 
    item_index=0, item_state_index=0))
    
# For each vehicle in Ego and Agent vehicles list, 
# record x,y positions, orientation and speed:
for vhc_ind in range(len(sim_environment.ego_vehicles_list) + len(sim_environment.agent_vehicles_list)):
  sim_environment.data_log_description_list.append(
    ItemDescription(item_type=ItemDescription.ITEM_TYPE_VEHICLE,
      item_index=vhc_ind,
      item_state_index=WebotsVehicle.STATE_ID_POSITION_X))
  sim_environment.data_log_description_list.append(
    ItemDescription(item_type=ItemDescription.ITEM_TYPE_VEHICLE,
      item_index=vhc_ind,
      item_state_index=WebotsVehicle.STATE_ID_POSITION_Y))
  sim_environment.data_log_description_list.append(
    ItemDescription(item_type=ItemDescription.ITEM_TYPE_VEHICLE,
      item_index=vhc_ind,
      item_state_index=WebotsVehicle.STATE_ID_ORIENTATION))
  sim_environment.data_log_description_list.append(
    ItemDescription(item_type=ItemDescription.ITEM_TYPE_VEHICLE,
      item_index=vhc_ind,
      item_state_index=WebotsVehicle.STATE_ID_SPEED))

# For each pedestrian, record x and y positions:
for ped_ind in range(len(sim_environment.pedestrians_list)):
  sim_environment.data_log_description_list.append(
    ItemDescription(item_type=ItemDescription.ITEM_TYPE_PEDESTRIAN,
      item_index=ped_ind,
      item_state_index=WebotsVehicle.STATE_ID_POSITION_X))
  sim_environment.data_log_description_list.append(
    ItemDescription(item_type=ItemDescription.ITEM_TYPE_PEDESTRIAN,
      item_index=ped_ind,
      item_state_index=WebotsVehicle.STATE_ID_POSITION_Y))

# Set the period of data log collection from the simulation.
sim_environment.data_log_period_ms = 10

# Create Trajectory dictionary for later reference.
# Dictionary will be used to relate received simulation trace to object states.
sim_environment.populate_simulation_trace_dict()
\end{lstlisting}
\end{exmp}

\subsubsection{Initial State Configuration}
Initial state configuration objects are for setting an initial state ofa simulation entity.

\begin{table}[H]
	\begin{center}
		\caption{Initial State Configuration.}
		\label{tab:initialstateconfigclass}
		\begin{tabularx}{\textwidth}{p{2.2cm}p{2.2cm}X}
			\hline
			\multicolumn{3}{l}{Class Name: \textbf{InitialStateConfig}}\\\hline
			\textbf{Property} & \textbf{Default} & \textbf{Description}\\\hline
			item & None & Data item for the corresponding state as an \textbf{ItemDescription} object.\\
			value & None & Initial value of the corresponding state.\\\hline
		\end{tabularx}
	\end{center}
\end{table}

\begin{exmp}
An example initial state configuration.
\begin{lstlisting}[language=Python, caption=Source code for setting an initial state value.]
from Sim_ATAV.simulation_control.initial_state_config import InitialStateConfig
from Sim_ATAV.simulation_control.item_description import ItemDescription

ego_init_speed_m_s = 10.0  # Keeping the Ego initial speed in a variable

# Create and add an initial state configuration into simulation environment object.
sim_environment.initial_state_config_list.append(
  InitialStateConfig(item=ItemDescription(
    item_type=ItemDescription.ITEM_TYPE_VEHICLE,  # State of a vehicle
    item_index=0,  # Vehicle index 0 (first added vehicle)
    item_state_index=WebotsVehicle.STATE_ID_VELOCITY_X),  # Speed along x-axis
    value=ego_init_speed_m_s))
\end{lstlisting}
\end{exmp}

\subsubsection{Viewpoint Configuration}
Webots viewpoint can automatically follow a simulation entity throughout the simulation.
Viewpoint configuration is used to describe which object to follow if desired.

\begin{table}[H]
	\begin{center}
		\caption{Viewpoint Configuration.}
		\label{tab:viewfollowconfigclass}
		\begin{tabularx}{\textwidth}{p{3.5cm}p{3cm}X}
			\hline
			\multicolumn{3}{l}{Class Name: \textbf{ViewFollowConfig}}\\\hline
			\textbf{Property} & \textbf{Default} & \textbf{Description}\\\hline
			item\_type & None & Type of the corresponding simulation entity. \\
			& & $<$TIME, VEHICLE, PEDESTRIAN$>$\\
			item\_index & None & Index of the corresponding simulation entity.\\
			position & None & $[x,y,z]$ values of the initial position of the viewpoint. \\
			rotation & None & $[x,y,z,\theta]$ Rotation of the viewpoint. \\\hline
		\end{tabularx}
	\end{center}
\end{table}

\begin{exmp}
An example viewpoint configuration.
\begin{lstlisting}[language=Python, caption=Source code for creating a viewpoint configuration.]
from Sim_ATAV.simulation_configurator.view_follow_config import ViewFollowConfig

# Create a viewpoint configuration to follow Ego vehicle (vehicle indexed as 0 as it 
# is the first added vehicle). Viewpoint will be positioned 15m behind and 3m above 
# the vehicle.
sim_environment.view_follow_config = \
  ViewFollowConfig(item_type=ItemDescription.ITEM_TYPE_VEHICLE,
    item_index=0,
    position=[sim_environment.ego_vehicles_list[0].current_position[0] - 15.0,
    sim_environment.ego_vehicles_list[0].current_position[1] + 3.0,
    sim_environment.ego_vehicles_list[0].current_position[2]],
    rotation=[0.0, 1.0, 0.0, \
      -sim_environment.ego_vehicles_list[0].current_orientation])
\end{lstlisting}
\end{exmp}

\subsubsection{Simulation Configuration}
Simulation configuration provides the information necessary to execute a simulation through TCP/IP connection between \toolblock{Simulation Configurator} and \toolblock{Simulation Supervisor}.
\begin{table}[H]
	\begin{center}
		\caption{Simulation Configuration.}
		\label{tab:simconfigclass}
		\begin{tabularx}{\textwidth}{llX}
			\hline
			\multicolumn{3}{l}{Class Name: \textbf{SimulationConfig}}\\\hline
			\textbf{Property} & \textbf{Default} & \textbf{Description}\\\hline
			world\_file & ``../Webots\_Projects/ & Name of the base Webots \\
			& worlds/test\_world\_1.wbt'' & world file.\\
			server\_port & 10021 & Port number for connecting to \toolblock{Simulation Supervisor}.\\
			server\_ip & ``127.0.0.1'' & Port number for connecting to \toolblock{Simulation Supervisor}.\\
			sim\_duration\_ms & 50000 & Simulation duration (ms).\\
			sim\_step\_size & 10 & Simulation time step (ms).\\
			run\_config\_arr& [] & List of run configurations for supporting multiple parallel simulation executions (\textbf{RunConfig} objects).\\\hline
			\multicolumn{3}{l}{Class Name: \textbf{RunConfig}}\\\hline
			\textbf{Property} & \textbf{Default} & \textbf{Description}\\\hline
			simulation\_run\_mode & FAST\_NO\_GRAPHICS & Webots run mode (simulation speed). \\
			& & REAL\_TIME: \\
			& & Real-time speed, \\
			& & FAST\_RUN: \\
			& & As fast as possible with visualization, \\
			& & FAST\_NO\_GRAPHICS: \\
			& & As fast as possible without visualization.\\\hline
		\end{tabularx}
	\end{center}
\end{table}

\begin{exmp}
An example simulation configuration creation.
\begin{lstlisting}[language=Python, caption=Source code for creating a simulation configuration.]
from Sim_ATAV.simulation_configurator import sim_config_tools

sim_config = sim_config_tools.SimulationConfig(1)
sim_config.run_config_arr.append(sim_config_tools.RunConfig())
sim_config.run_config_arr[0].simulation_run_mode = SimData.SIM_TYPE_REAL_TIME
sim_config.sim_duration_ms = 15000  # 15s simulation
sim_config.sim_step_size = 10
sim_config.world_file = '../Webots_Projects/worlds/empty_world.wbt'
\end{lstlisting}
\end{exmp}

\subsection{Executing the Simulation}
To start execution of a scenario, it is advised to first start Webots with the base world file manually.
If Webots crashes or communication is lost, \toolname can restart Webots when necessary (only is Webots is executing in the same system).
To start the simulation, \toolblock{Simulation Configurator} should be used to connect to the \toolblock{Simulation Supervisor}, send the simulation environment and configuration details, start the simulation, and finally collect the simulation trace.
Below is an example source code for this step.

\begin{exmp}
An example for execution of the simulation.
\begin{lstlisting}[language=Python, caption=Source code for executing the scenario.]
from Sim_ATAV.simulation_configurator.sim_environment_configurator import SimEnvironmentConfigurator

# Create a Simulation Configurator with the previously defined sim. configuration.
sim_env_configurator = SimEnvironmentConfigurator(sim_config=sim_config)

# Connect to the Simulation Supervisor. 
# Try maximum of 3 (max_connection_retry) times to connect.
(is_connected, simulator_instance) = sim_env_configurator.connect(max_connection_retry=3)
if not is_connected:
    raise ValueError('Could not connect!')

# Setup the scenario with previously populated Simulation Environment object.
sim_env_configurator.setup_sim_environment(sim_environment)

# Execute the simulation and get the simulation trace.
trajectory = sim_env_configurator.run_simulation_get_trace()
\end{lstlisting}
\end{exmp}

\pagebreak
\section{Combinatorial Testing}
\toolname provides basic functionality to read test scenarios from \filecall{csv} files, in which, each row represents a test case and each column holds the values of a test parameter.
This feature can be used to automatically execute a set of predefined test cases.
The user can create csv files that contain a set of test cases either manually or by using a tool.
\tbl{\ref{tab:csvfilereadfunctions}} provides a list of methods provided in \filecall{covering\_array\_utilities.py}.

\begin{table}[H]
	\begin{center}
		\begin{tabularx}{\textwidth}{lX}
			\hline
			\multicolumn{2}{l}{\textbf{load\_experiment\_data}(file\_name, header\_line\_count=6, index\_col=None)}\\\hline
			\textit{Description:} & Loads test cases from a csv file.\\
			\textit{Arguments:} & file\_name: Name of the csv file\\
			& header\_line\_count: Number of header lines to skip. (default: 6)\\
			& index\_col: Index of the column that is used for data indexing. (default: None)\\
			\textit{Return:} & A \textit{pandas} data frame (a tabular data structure) containing all test cases.\\\hline
			\multicolumn{2}{l}{\textbf{get\_experiment\_all\_fields}(exp\_data\_frame, exp\_ind)}\\\hline
			\textit{Description:} & Returns all parameters for a test case (one row of the test table).\\
			\textit{Arguments:} & exp\_data\_frame: Data frame that was read using\\
			& \textbf{load\_experiment\_data} \\
			& exp\_ind: Index of the experiment\\
			\textit{Return:} & A row from the test table that contains the requested test case.\\\hline
			\multicolumn{2}{l}{\textbf{get\_field\_value\_for\_current\_experiment}(cur\_experiment, field\_name)}\\\hline
			\textit{Description:} & Returns the value of the requested test parameter.\\
			\textit{Arguments:} & cur\_experiment: Current test case that was read using\\
			& \textbf{get\_experiment\_all\_fields}.\\
			& field\_name: Name of the test parameter.\\
			\textit{Return:} & Value of the requested parameter for the given test case.\\\hline
		\end{tabularx}
	\end{center}
	\caption{A list of Sim-ATAV methods for loading test cases from csv files.}
	\label{tab:csvfilereadfunctions}
\end{table}

For combinatorial testing, ACTS from NIST \cite{kuhn2013introduction} can be used as a tool for generating covering arrays.
Providing a complete guide on covering arrays and how to use ACTS is out of the scope of this chapter.
The user of ACTS creates a system definition by defining the system parameters and providing the possible values for each parameter.
The system definition is saved as an \filecall{xml} file.
ACTS can generate a covering array of desired strength and export the outputs in a \filecall{csv} file.
An example of a system definition and a corresponding 2-way covering array are provided in \toolname distribution in the files \filecall{tests/examples/TutorialExampleSystemACTS.xml} and \\ \filecall{tests/examples/TutorialExample\_CA\_2way.csv}, respectively.
Below is an example of reading test cases from a csv file and running each test one by one.
Note that, in the example, simulation trajectories are not evaluated. 
Outputs of each test case should be evaluated to decide the test result.

\begin{exmp}
	An example for running covering array test cases.
	
	In this example, a 2-way covering array for the test parameters is generated in ACTS.
	\tbl{\ref{tab:ca_test}} shows the original test parameter name from the file  \filecall{tutorial\_example\_1.py}, the name used in ACTS, and the possible values for each parameter (corresponding ACTS file is \filecall{TutorialExampleSystemACTS.xml}).
	\begin{table}[H]
		\begin{center}
			\caption{Describing test parameters for creating a covering array.}
			\label{tab:ca_test}
			\begin{tabularx}{\textwidth}{|X|X|X|}
				\hline
				\textbf{Parameter} & \textbf{Name in ACTS} & \textbf{Possible Values} \\\hline
				ego\_init\_speed\_m\_s & ego\_init\_speed & 0, 5, 10, 15 \\\hline
				ego\_x\_pos & ego\_x\_position & 15, 20, 25 \\\hline
				pedestrian\_speed & pedestrian\_speed & 2,3,4,5\\\hline
			\end{tabularx}
		\end{center}
	\end{table}

After entering the parameter descriptions in \tbl{\ref{tab:ca_test}} to ACTS, a 2-way covering array is generated and exported as a csv file.
Listing \ref{lst:csvlisting} shows the contents of the output csv file.
The file is available as: \filecall{tests/examples/TutorialExample\_CA\_2way.csv}.

\begin{lstlisting}[caption=CSV file containing a set of test cases generated by ACTS., label=lst:csvlisting, backgroundcolor=\color{gray!5},numbers=none]
# ACTS Test Suite Generation: Mon Jan 14 22:46:33 MST 2019
#  '*' represents don't care value 
# Degree of interaction coverage: 2
# Number of parameters: 3
# Maximum number of values per parameter: 4
# Number of configurations: 16
ego_init_speed,ego_x_position,pedestrian_speed
0,20,2
0,25,3
0,15,4
0,20,5
5,25,2
5,15,3
5,20,4
5,25,5
10,15,2
10,20,3
10,25,4
10,15,5
15,20,2
15,25,3
15,15,4
15,15,5
\end{lstlisting}

Below is the source code that is reading the test cases from the csv file using the functions listed in \tbl{\ref{tab:csvfilereadfunctions}} and running each test case one by one.
	
\begin{lstlisting}[language=Python, caption=Source code for executing the covering array test cases.]
import time
from Sim_ATAV.simulation_configurator import covering_array_utilities

def run_test(ego_init_speed_m_s=10.0, ego_x_pos=20.0, pedestrian_speed=3.0):
    """Runs a test with the given arguments"""
    .
    .
    .
    # This function is creating the test scenario, executing it and returning trajectory as described in previous section. Content is not repeated here for space considerations.
    .
    .
    .
    return trajectory
  
def run_covering_array_tests():
    """Runs all tests from the covering array csv file"""
    exp_file_name = 'TutorialExample_CA_2way.csv'   # csv file containing the tests
    
    # Read all experiment into a table:
    exp_data_frame = covering_array_utilities.load_experiment_data(exp_file_name, header_line_count=6)
    
    # Decide number of experiments based on the number of entries in the table.
    num_of_experiments = len(exp_data_frame.index)
    
    trajectories_dict = {}  # A dictionary data structure to keep simulation traces.
    
    for exp_ind in range(num_of_experiments):  # For each test case
        # Read the current test case
        current_experiment = covering_array_utilities.get_experiment_all_fields(
          exp_data_frame, exp_ind)
        
        # Read the parameters from the current test case:
        ego_init_speed = float(
          covering_array_utilities.get_field_value_for_current_experiment(
            current_experiment, 'ego_init_speed'))
        ego_x_position = float(
          covering_array_utilities.get_field_value_for_current_experiment(
            current_experiment, 'ego_x_position'))
        pedestrian_speed = float(
          covering_array_utilities.get_field_value_for_current_experiment(
            current_experiment, 'pedestrian_speed'))
        
        # Execute the test case and record the resulting simulation trace:
        trajectories_dict[exp_ind] = run_test(ego_init_speed_m_s=ego_init_speed, 
          ego_x_pos=ego_x_position, pedestrian_speed=pedestrian_speed)
        time.delay(2)  # Give Webots some time to reload the world
    return trajectories_dict
\end{lstlisting}
\end{exmp}

\section{Falsification / Search-based Testing}
For performing falsification, a CPS falsification tool like \staliro \cite{FainekosSUY12acc} can be used.
As \staliro is in \matlab, we need to call the \toolname test cases which are developed in \python from \matlab.
In this section, first a simple approach is described to call the test cases from \matlab, then a simple \staliro setup is described to perform falsification.
This chapter does not provide a complete guide to \staliro. 
Reader is referred to \staliro website for further details \footnote{\staliro iwebsite: https://sites.google.com/a/asu.edu/s-taliro/s-taliro}. 

\subsection{Connecting to \matlab}
\matlabCopyrighted has built-in support to call \python functions.
However, type conversions are required both in \toolname and in the \matlab code \footnote{The user is free to use a different approach than what is described here.}.
\toolname provides basic functionality to convert simulation trace to a format (single dimensional list) that can be easily interpreted in \matlab.
Listing \ref{lst:pythoncodeformatlab} shows the necessary updates on the \python side.
First, \filecall{sim\_duration} argument is added to \filecall{run\_test} function allow \staliro run different length of simulations.
Next, \filecall{for\_matlab} argument is added to tell the function is called from \matlab so that \filecall{experiment\_tools.npArray2Matlab} method from \toolname is used to convert simulation trace to a single list for \matlab.

\begin{lstlisting}[language=Python, caption=Source code for executing the covering array test cases., label=lst:pythoncodeformatlab]
def run_test(ego_init_speed_m_s=10.0, ego_x_pos=20.0, pedestrian_speed=3.0, sim_duration=15000, for_matlab=False):
    """Runs a test with the given arguments"""
    ...
    # This function is creating the test scenario, executing it and returning trajectory as described in previous section. Content is not repeated here for space considerations.
    ...
    if for_matlab:
        trajectory = experiment_tools.npArray2Matlab(trajectory)
    return trajectory
\end{lstlisting}

On the \matlab side, we need a wrapper function to call the function \filecall{run\_test}.
Listing \ref{lst:matlabcodeforpython} provides an example to such a wrapper function.
This \matlab code can be found in \toolname distribution in the file \filecall{/tests/examples/run\_tutorial\_example\_from\_matlab.m}.
Unused return parameters \filecall{YT, LT, CLG, GRD} and input arguments \filecall{steptime, InpSignal} are to maintain compatibility with \staliro function call format.

\begin{lstlisting}[language=Matlab, caption=Matlab code as a wrapper to execute Python test execution function., label=lst:matlabcodeforpython,backgroundcolor=\color{blue!3}]
function [T, XT, YT, LT, CLG, GRD] = run_tutorial_example_from_matlab(XPoint, sim_duration_s, steptime, InpSignal)
%run_tutorial_example_from_matlab Run Webots simulation with the parameters in XPoint.
% XPoint contains: [ego_init_speed_m_s, ego_x_pos, pedestrian_speed]

% Run the simulation and receive the trajectory:
traj = py.tutorial_example_1.run_test(XPoint(1), XPoint(2), XPoint(3), int32(sim_duration_s*1000.0), true);

% Convert trajectory to matlab array
mattraj = Core_py2matlab(traj);  % Core_py2matlab is from Matlab fileexchange, developed by Kyle Wayne Karhohs
YT = [];
LT = [];
CLG = [];
GRD = [];
if isempty(mattraj)
  T = [];
  XT = [];
else
  % Separate time from the simulation trace:
  T = mattraj(:,1)/1000.0;  % Also, convert time to s from ms
  XT = mattraj(:, 2:end);  % Rest of the trace
end
end
\end{lstlisting}

\subsection{Connecting to \staliro}
To perform falsification with \staliro, an MTL specification should be defined, ranges for test parameters and the \matlab code, which is given in Listing \ref{lst:matlabcodeforpython}, for running a simulation should be provided to \staliro as the model under test.
Listing \ref{lst:stalirocode} gives a very simple example for falsification with an MTL requirement that checks x and y coordinates of Ego and agent vehicles to decide a collision.
This \matlab code can be found in \toolname distribution in \filecall{/tests/examples/run\_falsification.m}.
The requirement is not to collide onto the agent vehicle.
Note that, this requirement is very simplified to provide a simple example source code.
A more complete check for collision would require incorporating further vehicle details and/or extracting collision information from the simulation.

\begin{lstlisting}[language=Matlab, caption=Matlab code for running falsification with S-TaLiRo., label=lst:stalirocode, ,backgroundcolor=\color{blue!3}]
% Indices of states in simulation trace:
cur_traj_ind = 1;
EGO_X = cur_traj_ind; cur_traj_ind = cur_traj_ind + 1;
EGO_Y = cur_traj_ind; cur_traj_ind = cur_traj_ind + 1;
EGO_THETA = cur_traj_ind; cur_traj_ind = cur_traj_ind + 1;
EGO_V = cur_traj_ind; cur_traj_ind = cur_traj_ind + 1;
AGENT_X = cur_traj_ind; cur_traj_ind = cur_traj_ind + 1;
AGENT_Y = cur_traj_ind; cur_traj_ind = cur_traj_ind + 1;
AGENT_THETA = cur_traj_ind; cur_traj_ind = cur_traj_ind + 1;
AGENT_V = cur_traj_ind; cur_traj_ind = cur_traj_ind + 1;
PED_X = cur_traj_ind; cur_traj_ind = cur_traj_ind + 1;
PED_Y = cur_traj_ind;
NUM_ITEMS_IN_TRAJ = cur_traj_ind;

% Predicates for MTL requirement:
ii = 1;
preds(ii).str='y_check1';
preds(ii).A = zeros(1, NUM_ITEMS_IN_TRAJ);
preds(ii).A(AGENT_Y) = 1;
preds(ii).A(EGO_Y) = -1;
preds(ii).b = 1.5;

ii = ii+1;
preds(ii).str='y_check2';
preds(ii).A = zeros(1, NUM_ITEMS_IN_TRAJ);
preds(ii).A(AGENT_Y) = -1;
preds(ii).A(EGO_Y) = 1;
preds(ii).b = 1.5;

ii = ii+1;
preds(ii).str='x_check1';
preds(ii).A = zeros(1, NUM_ITEMS_IN_TRAJ);
preds(ii).A(AGENT_X) = 1;
preds(ii).A(EGO_X) = -1;
preds(ii).b = 8;

ii = ii+1;
preds(ii).str='x_check2';
preds(ii).A = zeros(1, NUM_ITEMS_IN_TRAJ);
preds(ii).A(AGENT_X) = -1;
preds(ii).A(EGO_X) = 1;
preds(ii).b = 0;

% Metric Temporal Logic Requirement:
phi = '[](!(y_check1 /\ y_check2 /\ x_check1 /\ x_check2))';

% Ranges for test parameters (ego_init_speed_m_s, ego_x_pos, pedestrian_speed):
init_cond = [0.0, 15.0;
            15.0, 25.0;
             2.0, 5.0];

% Provide our Matlab wrapper function for running the tests as the model.
model = @run_tutorial_example_from_matlab;
opt = staliro_options();
opt.runs = 1;  % Do falsification only once.
opt.black_box = 1;  % Because we use a custom Matlab function as the model.
opt.SampTime = 0.010;  % Sample time. Same as Webots world time step.
opt.spec_space = 'X';  % Requirements are defined on state space.
opt.optimization_solver = 'SA_Taliro';  % Use Simulated Annealing
opt.taliro = 'dp_taliro';  % Use dp_taliro to compute robustness
opt.map2line = 0;
opt.falsification = 1;  % Stop when falsified
opt.optim_params.n_tests = 100;  % maximum number of tries
sim_duration = 15.0;

disp(['Running S-TaLiRo ... '])
[results, history] = staliro(model, init_cond, [], [], phi, preds, sim_duration, opt);

res_filename = ['results_', datestr(datetime("now"), 'yyyy_mm_dd__HH_MM'), '.mat'];
save(res_filename)
disp(["Results are saved to: ", res_filename])
\end{lstlisting}

\section{Other Remarks}
A user guide for \toolname is provided with a running example.
The source code for the running example used in this chapter is available in \toolname distribution under \filecall{tests/examples} folder.
As \toolname is a research tool that is not directly targeted for production-level systems special caution should be taken before using it for testing any critical functionality.
\toolname is still evolving and it may contain a number of bugs or parts that are open to optimization.
Here, only major functionality provided by \toolname is discussed.
It provides further functionality like a number of sample vehicle controller subsystems, additional computations on simulation trajectory such as collision detection, and functionality toward evaluating perception-system performance.
For a deeper understanding of \toolname's capabilities, the reader is encouraged to go through the source code for examples and experiments that are used as case studies for publications.


\bibliographystyle{eptcs}
\bibliography{rrtbased_bib.bib}

\end{document}